\documentclass[letterpaper, 10 pt, conference]{ieeeconf}  

\IEEEoverridecommandlockouts                              

\usepackage{graphics} 
\usepackage{graphicx} 
\usepackage{epsfig} 
\usepackage{amsmath, bm} 
\usepackage{amssymb}  
\usepackage{upgreek}
\usepackage{multicol}
\usepackage{multirow}
\usepackage{lipsum}  
\usepackage{breqn}
\usepackage{comment} 
\usepackage{booktabs}

\PassOptionsToPackage{hyphens}{url}\usepackage{hyperref}

\usepackage{xcolor}

\usepackage[style=ieee]{biblatex}

\DeclareSourcemap{
  \maps{
    \map{
      \pertype{article}
      \step[fieldset=language, null]
      \step[fieldset=url, null]
      \step[fieldset=doi, null]
      \step[fieldset=issn, null]
      \step[fieldset=isbn, null]
      \step[fieldset=note, null]
      \step[fieldset=editor, null]
      \step[fieldset=urldate, null]
      \step[fieldset=file, null]
    }
  }
}
\DeclareSourcemap{
  \maps{
    \map{
      \pertype{inproceedings}
      \step[fieldset=language, null]
      \step[fieldset=url, null]
      \step[fieldset=doi, null]
      \step[fieldset=issn, null]
      \step[fieldset=isbn, null]
      \step[fieldset=note, null]
      \step[fieldset=editor, null]
      \step[fieldset=urldate, null]
      \step[fieldset=file, null]
    }
  }
}
\DeclareSourcemap{
  \maps{
    \map{
      \pertype{incollection}
      \step[fieldset=language, null]
      \step[fieldset=url, null]
      \step[fieldset=doi, null]
      \step[fieldset=issn, null]
      \step[fieldset=isbn, null]
      \step[fieldset=note, null]
      \step[fieldset=editor, null]
      \step[fieldset=urldate, null]
      \step[fieldset=file, null]
    }
  }
}

\title{\LARGE \bf
Bang-Bang Control Of A Tail-less Morphing Wing Flight 
}

\author{Eric Sihite$^{1}$ $^{\dagger}$, Xintao Hu$^{2}$ $^{\dagger}$, Bozhen Li$^{2}$, Adarsh Salagame$^{2}$, Paul Ghanem$^{2}$, and Alireza Ramezani$^{2}$%
\thanks{$^{\dagger}$ These authors have equal contribution to this paper.}%
\thanks{$^{1}$ The author is with the Department of Aerospace, California Institute of Technology, Pasadena, CA-91125, USA. (e-mail: esihite@caltech.edu).}%
\thanks{$^{2}$ The author is with the SiliconSynapse Laboratory, Department of Electrical and Computer Engineering, Northeastern University, Boston, MA-02119, USA. (e-mail: hu.xinta, li.bozh, salagame.a, ghanem.p, a.ramezani@northeastern.edu)}%
}

\begin{document}

\maketitle
\thispagestyle{empty}
\pagestyle{empty}

\begin{abstract}

Bats' dynamic morphing wings are known to be extremely high-dimensional, and they employ the combination of inertial dynamics and aerodynamics manipulations to showcase extremely agile maneuvers. Bats heavily rely on their highly flexible wings and are capable of dynamically morphing their wings to adjust aerodynamic and inertial forces applied to their wing and perform sharp banking turns. There are technical hardware and control challenges in copying the morphing wing flight capabilities of flying animals. This work is majorly focused on the modeling and control aspects of stable, tail-less, morphing wing flight. A classical control approach using bang-bang control is proposed to stabilize a bio-inspired morphing wing robot called \textit{Aerobat}. Robot-environment interactions based on horseshoe vortex shedding and Wagner functions is derived to realistically evaluate the feasibility of the bang-bang control, which is then implemented on the robot in experiments to demonstrate first-time closed-loop stable flights of Aerobat.

\end{abstract}


\section{Introduction}
\label{sec:introduction}

Bats' dynamic morphing wings are known to be extremely high-dimensional, involving the synchronous movements of many active and passive coordinates, joint clusters, in a gait cycle. These animals apply their unique array of specializations to dynamically morph the shape of their wings to enhance their agility and energy efficiency. Copying bat dynamic morphing wing can bring fresh perspectives to micro aerial vehicle (MAV) design \cite{ramezani_biomimetic_2017}.

For instance, bats employ the combination of inertial dynamics and aerodynamics manipulations to showcase extremely agile maneuvers. Unlike rotary- and fixed-wing systems wherein aerodynamic surfaces (e.g., ailerons, rudders, propellers, etc.) come with the sole role of aerodynamic force adjustments, the articulated wings in bats possess more sophisticated roles \cite{riskin_bats_2009}. Or, it is known that bats can perform zero-angular-momentum turns by making differential adjustments (e.g., collapsing armwings) in the inertial forces led by their wings. Bats can apply a similar mechanism to perform sharp banking turns \cite{riskin_upstroke_2012,iriarte-diaz_whole-body_2011}. 

Several attempts have been made to copy flapping flight of animals including insects, birds, bats, etc., ranging from smaller insect-sized robots, or micro UAVs \cite{phan_insect-inspired_2019, farrell_helbling_review_2018, ma_controlled_2013, chukewad_robofly_2020, tu_untethered_2020, rosen_development_2016}, to bat or small bird-sized robots with a wingspan between 20 and 60 cm \cite{hoff_synergistic_2016, ramezani_bat_2016, ramezani_biomimetic_2017, hoff_optimizing_2018, sihite_computational_2020, de_croon_design_2009, peterson_wing-assisted_2011, wissa_free_2015}, and larger robots with wingspan larger than 1 m \cite{send_artificial_2012, gerdes_robo_2014}. Unfortunately, most of these examples fail to copy dynamic morphing capabilities manifested by the powered flight of animals.

\begin{figure}[t]
    \centering
    \vspace{0.1in}
    \includegraphics[width = \linewidth]{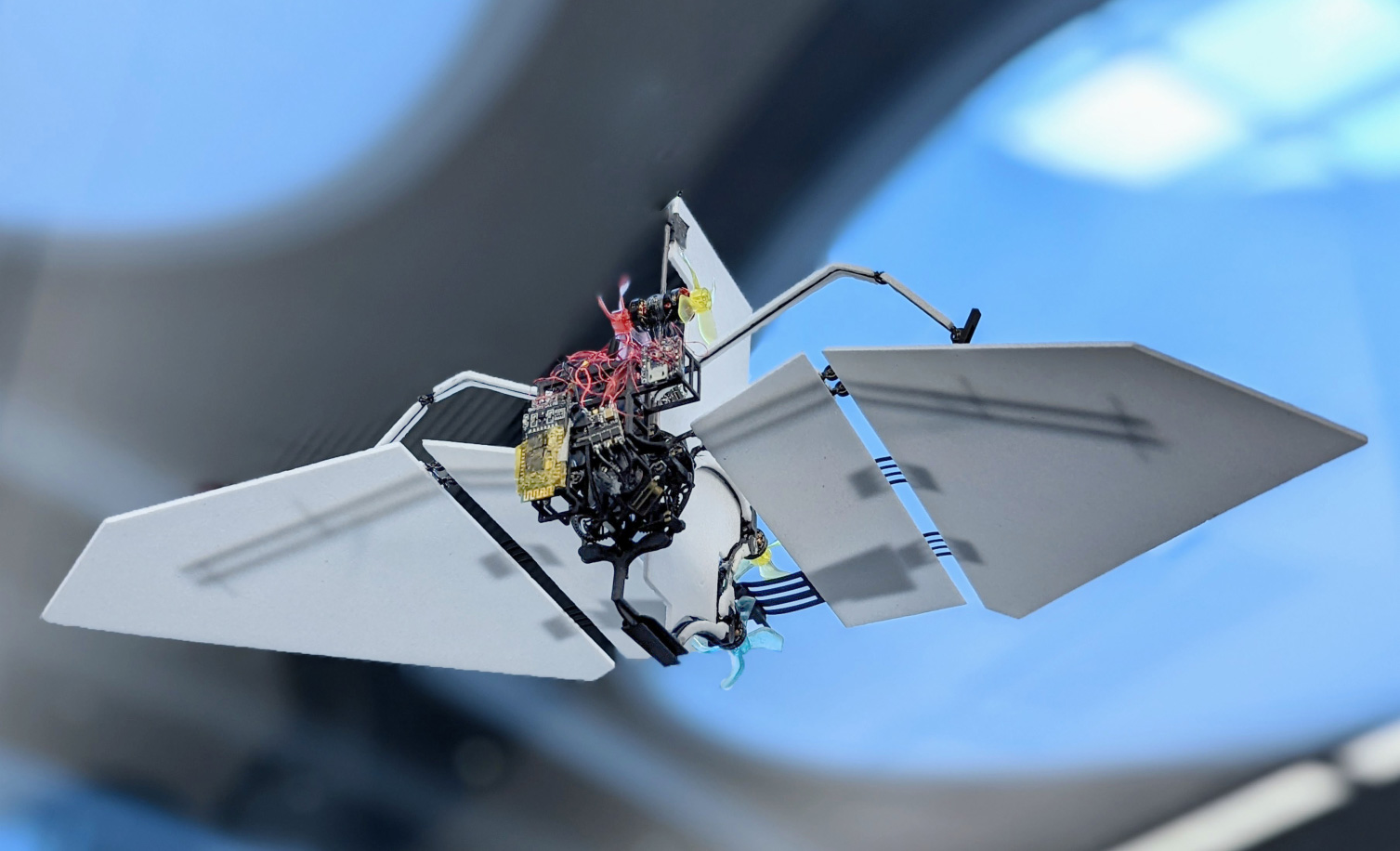}
    \caption{Shows Northeastern University's morphing wing robot, \textit{Aerobat}. Aerobat is employed to test our bang-bang flight control.}
    \vspace{-0.1in}
    \label{fig:cover}
\end{figure}

Other than technical hardware challenges facing copying morphing wing flight there are modeling and control challenges. This work is majorly focused on the modeling and control aspects of morphing wing flight. We propose a classical bang-bang control approach tested on a robot called \textit{Aerobat} (see Fig.~\ref{fig:cover}) being developed at Northeastern University (NU). While our proposed method remains classical and not novel, we had to overcome a number of technical challenges worthy of writing this report to demonstrate the closed-loop flight of Aerobat in an untethered fashion for the first time.

Control of dynamic morphing wing flight is an extremely challenging problem \cite{ramezani_lagrangian_2015, mangan_describing_2017, mangan_reducing_2017, hoff_trajectory_2019, sihite_enforcing_2020}. As part of our past work, we have developed simulation models to investigate control methods \cite{sihite_enforcing_2020}, and incorporating embodied locomotion through a change in morphology using mechanical intelligence \cite{sihite_integrated_2021}. We explored models of robot-environment interaction based on Dickinson's celebrated work \cite{sane_lift_2001}. However, so far these models fail to capture leading- or trailing-edge vortex shedding \cite{hedenstrom_bat_2015, hubel_wake_2010} effects which are known to be pronounced in morphing wing flight. This research gap has motivated us to develop a more accurate model for our robot and exploit it to evaluate the feasibility of flight control based on bang-bang control in simulation and experiment.

This paper is outlined as follows: a brief discussion on the Aerobat platform featured in this paper, followed by the dynamic modeling, controller definition, and the numerical simulation to show the proof-of-concept of the bang-bang controller, then followed by experimental results of flying the robot using said controller, and concluding remarks.

\section{Aerobat Platform: A Platform to Study Dynamic Morphing Wing Flight}
\label{sec:hardware}

NU's Aerobat is a tail-less flapping robot that unlike existing examples are capable of significantly morphing wing structure dynamically during each gait cycle which is a fraction of a second. This robot, which weighs roughly 50-60 grams depending on the onboard sensors, with a wingspan of approximately 30 cm, was developed to study the flapping-wing flight of bats. 

Aerobat utilizes a computational structure, called the \textit{Kinetic Sculpture} (KS) \cite{sihite_computational_2020}, that introduces computational resources for wing morphing. The KS is designed to actuate the robot's wings as it is split into two wing segments: the proximal and distal wings, which are actuated by what is the equivalent of shoulder and elbow joints, respectively. The gait captures the wing folding during the upstroke motion, which is one of the key modes in bat flight. The wing folding reduces the wing surface area and minimizes the negative lift during the upstroke and results in a more efficient flight. Aerobat is capable of flapping at a frequency of up to 8 Hz using its onboard electronics.

This tail-less robot is unstable in its longitudinal (pitch dynamics) and frontal (roll dynamics) planes of flight. Therefore, it is necessary to develop a method to stabilize longitudinal and frontal dynamics. As part of our efforts to gradually move towards stable flight, in this work we explore closed-loop stabilization of the robot using a bang-bang control concept. The actuators considered for these tests constitute small thrusters that only deliver correction moments around the center of mass of Aerobat.


\section{Numerical Simulation}
\label{sec:simulation}

This section outlines the dynamical modeling and simulation utilized as a proof-of-concept for our bang-bang control method. The dynamic modeling is derived using an unsteady aerodynamic model using the Wagner model and lifting-line theory \cite{boutet_unsteady_2018}. Then, a bang-bang controller is developed to control Aerobat and show that the system is capable of flying stably.

\subsection{Modeling Inertial and Aerodynamic Contributions}
\label{subsec:modeling}

\begin{figure*}[t]
    \centering
    \vspace{0.1in}
    \includegraphics[width = 0.8 \linewidth]{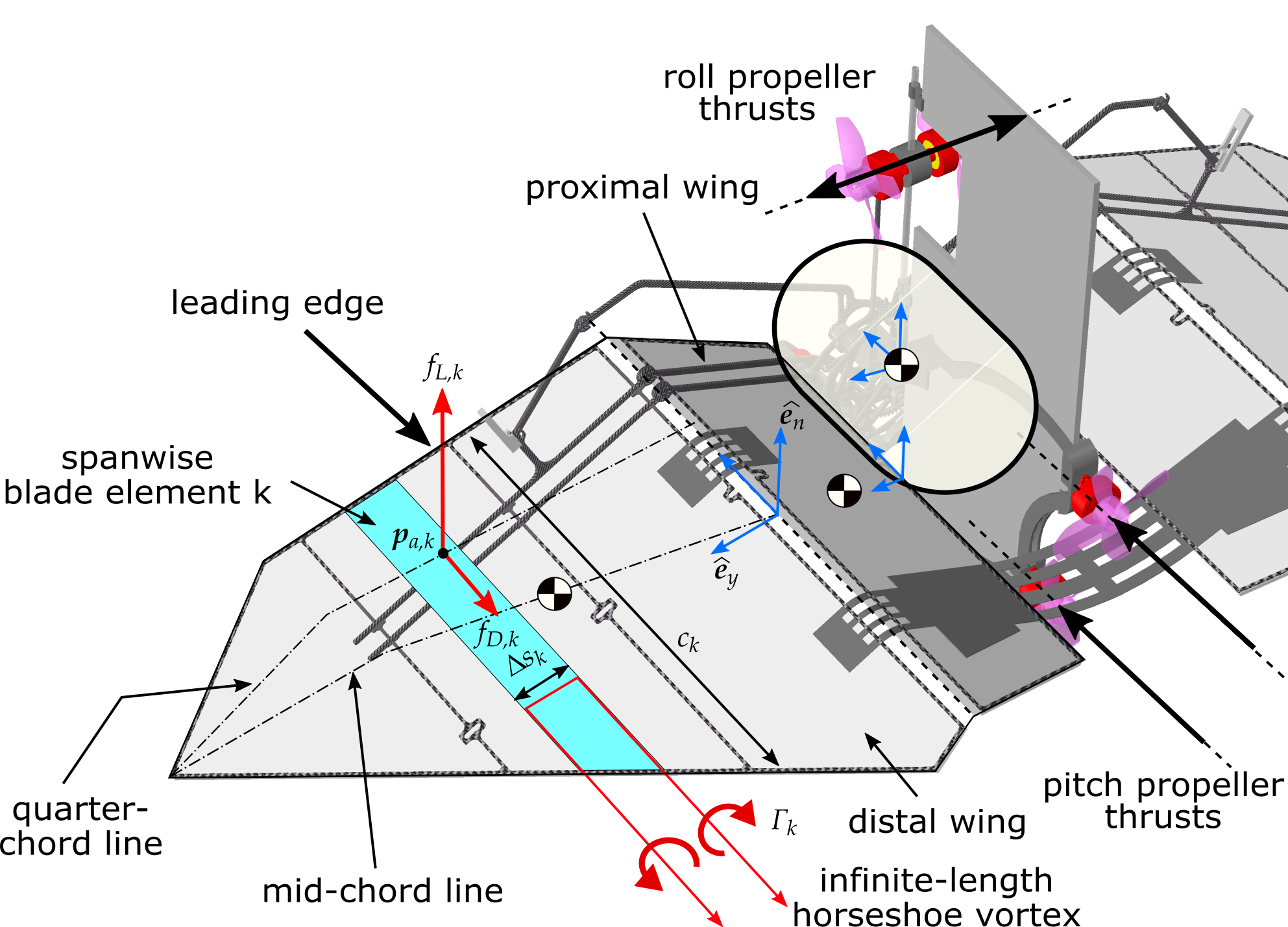}
    \caption{Illustrates how the vortex shedding is parameterized in the model of Aerobat. In addition, the small thrusters used to control longitudinal and frontal dynamics based on a bang-bang control concept are shown.}
    \vspace{-0.1in}
    \label{fig:fbd}
\end{figure*}

Aerobat has 20 degrees-of-freedom (DOF) present in the system which makes the dynamic modeling very difficult to derive. A simplification is performed to reduce the complexity of the simulation and computational time. The KS can be reduced using kinematics constraints down to a single DOF which is actuated by the motor. The joint trajectories corresponding to the shoulder and elbow joints can be used as a kinematic constraint for the dynamical components of the robot. The KS is also designed to synchronize the gait of both wings. This reduces the wing system down to 1 DOF represented by the motor angle. 

Including the body's 6 DOF, the simplified dynamical model of Aerobat can be represented with 7 DOF. The dynamical equation of motion used in the simulation can be derived using Euler-Lagrangian dynamical formulations. Figure \ref{fig:fbd} shows the free-body diagram of the robot, which can be presented using 5 bodies: main body, proximal and distal wings of both sides. The synchronized wing trajectory allows us to just use one side of the wing in the states.

Let $\bm q = [\bm p^\top, \bm \theta^\top, q_s, q_e]^\top$ be the generalized coordinates, where $\bm p$ is the body center of mass inertial position, $\bm \theta$ is the Euler angles of the body, $q_s$ and $q_e$ are the left wing's shoulder and elbow angles, respectively. The dynamical equation of motion of the simplified system can be defined as follows:
\begin{equation}
\begin{aligned}
    \bm M(\bm q) \, \ddot{\bm q} &= \bm h(\bm q, \dot{\bm q}) + \bm u_a + \bm u_t + \bm J_c^\top \bm \lambda  \\
    \bm J_c \, \ddot{\bm q} &= [\ddot q_s, \ddot q_e]^\top = \bm y_{ks}, 
\end{aligned}
\label{eq:dynamic_eom}
\end{equation}
where $M$ is the inertial matrix, $\bm h$ is the gravitational and Coriolis forces, $\bm u_a$ and $\bm u_t$ are the generalized aerodynamic and thruster forces, respectively. $\bm \lambda$ is the Lagrangian multiplier which enforces the constraint forces acting on $q_s$ and $q_e$ to track the KS flapping acceleration $\bm y_{ks}$. $\bm \lambda$ can be solved algebraically from \ref{eq:dynamic_eom} given the states $\bm x = [\bm q^\top, \dot{\bm q}^\top]^\top$ and both generalized forces $\bm u_a$ and $\bm u_t$. These generalized forces can be derived using virtual displacement, as follows:
\begin{equation}
\begin{aligned}
    \bm u_a &= \sum_{i=1}^{N_b} B_{a,i}(\bm q)\, \bm f_{a,i} \quad &
    \bm u_t &= \sum_{i=1}^{N_t} B_{t,i}(\bm q)\, \bm f_{t,i}
\end{aligned}
\label{eq:generalized_forces}
\end{equation}
where $B$ matrices map the forces $\bm f \in \mathbb^{R}^3$ to the generalized coordinates $\bm q$, $N_b$ is  the number of blade elements, and $N_b$ is the number of thrusters. Let the position $\bm p_k(\bm q)$ be the inertial position where the force $\bm f_k$ defined in the inertial frame is applied. The matrix $B_k$ for this force can be derived as follows: $B_k = \left( \partial \dot{\bm p}_{k} / \partial \dot{\bm q} \right)^\top$. The aerodynamic forces generated on each blade elements and thrust forces are combined to form $\bm u_a$ and $\bm u_t$, respectively.

The aerodynamics can be derived using discrete blade elements following the derivations in \cite{boutet_unsteady_2018}. This model uses the lifting line theory and Wagner's function to develop a model for calculating the lift coefficient. Let $S$ be the total wingspan and $y \in [-S/2, S/2]$ represents a position along the wingspan. The vortex shedding distribution can be defined as a function of truncated Fourier series of size $m$ across the wingspan, as follows:
\begin{equation}
\begin{gathered}
    \Gamma(t,y) = \frac{1}{2} a_0 \, c_0 \, U \, \sum^{m}_{n=1} a_n(t) \, \sin(n\,\theta(y))
\end{gathered}
\end{equation}
where $a_n$ is the Fourier coefficients, $a_0$ is the slope of the angle of attack, $c_0$ is the chord length at wing's axis of symmetry, and $U$ is the free stream airspeed. Let $\theta$ be the change of variable defined by $y = (S/2)\cos(\theta)$ for describing a position along the wingspan $y \in (-S/2, S/2)$. From $\Gamma(t,y)$, we can derive the additional downwash induced by the vortices, defined as follows:
\begin{equation}
\begin{aligned}
    w_{y}(t,y) &
    = - \frac{a_0 c_0 U}{4S} \sum^{m}_{n=1} n a_n(t)  \frac{\sin(n \theta)}{\sin(\theta)}.
\end{aligned}
\label{eq:induced_downwash}
\end{equation}
%

Following the unsteady Kutta-Joukowski theorem, the sectional lift coefficient can be expressed as follows:
\begin{equation}
\begin{aligned}
    C_L(t,y) &= a_0 \sum^{m}_{n=1} \left( \frac{c_0}{c(y)} a_n(t) + \frac{c_0}{U} \dot{a}_n(t) \right) \sin(n\theta),
\end{aligned}
\label{eq:lift_coeff_fourier}
\end{equation}
where $c(y)$ is the chord length at the wingspan position $y$. The computation of the sectional lift coefficient response of an airfoil undergoing a step change in downwash $\Delta w(y) << U$ can be expressed using Wagner function $\Phi(t)$:
\begin{equation}
\begin{aligned}
    c_L(t,y) &= \frac{a_0}{U} \Delta w(t,y) \Phi(\tilde t) \\
    \Phi(\tilde t)    &= 1 - \psi_1 e^{-\epsilon_1 \tilde t} - \psi_2 e^{-\epsilon_2 \tilde t}
\end{aligned}
\label{eq:lift_coeff_wagner}
\end{equation}
where $\tilde t(t) = \int_0^t (v_e^i/b) dt$ is the normalized time which is defined as the distance traveled divided by half chord length ($b = c/2$). Here, $v_e^i$ is defined as the velocity of the quarter chord distance from the leading edge in the direction perpendicular to the wing sweep. For the condition where the freestream airflow dominates $v_e$, then we can approximate the normalized time as $\tilde t = Ut/b$. The Wagner model in \eqref{eq:lift_coeff_wagner} uses Jones' approximation \cite{boutet_unsteady_2018}, with the following coefficients: $\psi_1 = 0.165$, $\psi_2 = 0.335$, $\epsilon_1 = 0.0455$, and $\epsilon_2 = 0.3$.

\begin{figure*}[t]
    \centering
    \vspace{0.1in}
    \includegraphics[width = \linewidth]{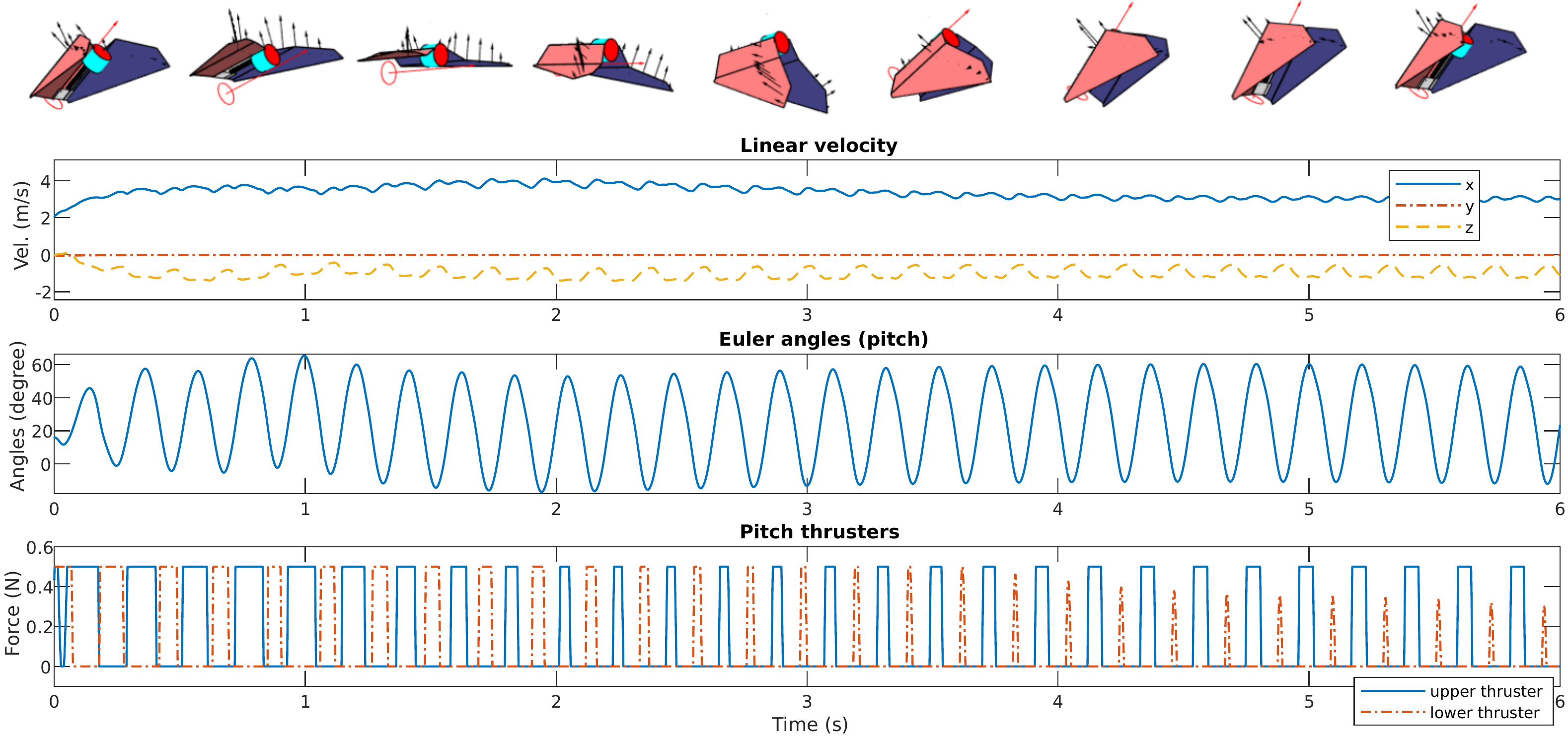}
    \caption{Illustrates Aerobat's stick-diagram and simulated state trajectories under bang-bang control of the longitudinal and frontal dynamics.}
    \vspace{-0.1in}
    \label{fig:simulation_results}
\end{figure*}

Duhamel's principles can be used to superimpose the transient response due to a step change in downwash as defined in \eqref{eq:lift_coeff_wagner}. Additionally, integration by parts can be used to simplify the equation further, resulting in the following equation:
\begin{equation}
\begin{aligned}
    C_L(t,y) &= \frac{a_0}{U} \left( w(t,y) \Phi(0) - \int_{0}^{t} \frac{\partial \Phi(t - \tau)}{\partial \tau} w(\tau, y) d\tau \right).
\end{aligned}
\label{eq:aero_CL_base}
\end{equation}
\begin{equation}
\begin{aligned}
    \frac{\partial \Phi(t - \tau)}{\partial \tau} &=
    -\frac{\psi_1 \epsilon_1 U}{b} e^{-\frac{\epsilon_1 U}{b}(t-\tau)}
    -\frac{\psi_2 \epsilon_2 U}{b} e^{-\frac{\epsilon_2 U}{b}(t-\tau)}
\end{aligned}
\label{eq:partial_phi}
\end{equation}
Here, $w(t,y)$ is the total downwash defined as:
\begin{equation}
    w(t,y) = v_n(t,y) + w_y(t,y),
\label{eq:total_downwash}
\end{equation}
where $v_n$ is the airfoil velocity normal to the wing surface which depends on the freestream velocity and the inertial dynamics. Finally, we can represent the integrals as the following states:
\begin{equation}
\begin{aligned}
    z_{1} (t,y) &= \int_{0}^{t} \frac{\psi_1 \epsilon_1 U}{b} e^{-\frac{\epsilon_1 U}{b}(t-\tau)} w(\tau,y) d\tau
    \\
    z_{2} (t,y) &= \int_{0}^{t} \frac{\psi_2 \epsilon_2 U}{b} e^{-\frac{\epsilon_2 
    U}{b}(t-\tau)} w(\tau,y) d\tau.
\end{aligned}
\label{eq:aero_states_z}
\end{equation}
Both of these states can be expressed as an ODE by deriving the time derivatives of \eqref{eq:aero_states_z}. They can be derived using Leibniz integral rule, yielding the following equations:
\begin{equation}
\begin{aligned}
    \dot z_{1} (t,y) &= \frac{\psi_1 \epsilon_1 U}{b} \left( w(t,y) - \frac{\epsilon_1 U}{b} z_1(t,y) \right) \\
    \dot z_{2} (t,y) &= \frac{\psi_2 \epsilon_2 U}{b} \left( w(t,y) - \frac{\epsilon_2 U}{b} z_2(t,y) \right).
\end{aligned}
\label{eq:aero_states_dz}
\end{equation}
The sectional lift coefficient can then be defined as:
\begin{equation}
\begin{aligned}
    c_L(t,y)  = \frac{a_0}{U} \left( w(t,y) \phi(0) + z_1(t,y) + z_2(t,y) \right),
\end{aligned}
\label{eq:aero_CL_final}
\end{equation}
and we can march the aerodynamic states $z_1$ and $z_2$ forward in time using \eqref{eq:aero_states_dz}. Finally, we can relate the both sectional lift coefficient equations in \eqref{eq:lift_coeff_fourier} and \eqref{eq:aero_CL_final} to solve for the Fourier coefficient rate of change, $\dot{a}_n$. 

The aerodynamic states are defined along the span of the wing and can be discretized into $m$ blade elements. Therefore, we can derive the $m$ equations relating \eqref{eq:lift_coeff_fourier} and \eqref{eq:aero_CL_final} on each blade element to solve for the $\dot{a}_n$. Then, including $z_1$ and $z_2$ on each blade elements, we will have $3m$ ODE equations to solve. We can represent $a_n$, $z_1$, and $z_2$ of all blade elements as the vector $\bm a_n \in \mathbb{R}^{m}$, $\bm {z}_1 \in \mathbb{R}^{m}$, and $\bm z_2 \in \mathbb{R}^{m}$, respectively. 


\subsection{Bang-Bang Control of Longitudinal and Frontal Dynamics}
\label{subsec:controller}

The orientation and speed control of the robot can be established using a change in the thruster forces and flapping speed. In this work, we attempt to keep the flapping speed constant at 4.75 Hz which is the flapping speed of the robot used in our experiments. Therefore, we only utilized the four thrusters to stabilize the robot's roll and pitch, in addition to it's forward speed. 

Let $v_i$ be the magnitude of the thrust forces of thruster $i$, as labeled in Fig. \ref{fig:fbd}. Let $v_1$ and $v_2$ be the backwards-facing thrusters above and below the robot, respectively, to adjust the robot's pitch and assist in thrust generation. Then, let $v_3$ and $v_4$ be the thrusters facing the robot's left and right, respectively, which are used to adjust the roll of Aerobat. The following bang-bang controller is used to stabilize the robot's orientation and forward speed:
\begin{equation}
\Gamma_{BB}=\left\{
\begin{aligned}
    v_1 = \text{ON} & \text{~~if~~} \theta_y > \theta_{y,ref} \\
    v_2 = \text{ON} & \text{~~if~~} \theta_y \leq \theta_{y,ref} \\
    v_3 = \text{ON} & \text{~~if~~} \theta_x > \theta_{x,ref} \\
    v_4 = \text{ON} & \text{~~if~~} \theta_x \leq \theta_{x,ref}, \\
\end{aligned}
\right.
\label{eq:controller}
\end{equation}
where $\theta_x$ and $\theta_y$ are the roll and pitch angles, respectively. We constraint the thrusters to only generate forces in one direction and assume that the torque produced by the propeller drag is negligible. 

\section{Results}
\label{sec:experiment}

Here, we briefly cover our simulation and experimental results from Aerobat's untethered flights in a straight path.

\subsection{Simulation Results and Discussions}
\label{subsec:simulation_result}

The simulation was set up to match some of the flight conditions of the actual robot in our experiments. In the simulation, the robot was initialized with an initial forward speed of 2 m/s. No upstream flow is assumed in these simulations. We use the flapping speed of 4.75 Hz and initialize the robot with a small perturbation in initial stats. The robot orientation was initialized with an initial pitch and roll of $15^\circ$ and $-5^\circ$, respectively. We set the pitch and roll reference to $20^\circ$ and $0^\circ$, respectively.

In Fig.~\ref{fig:simulation_results}, simulated states and thrusters forces during the simulation are shown. Top figure in Fig.~\ref{fig:simulation_results} shows the stick-diagram of Aerobat. The model reaches a stable limit-cycle within approximately 5 seconds. The bang-bang controller despite its simplicity is capable of stabilizing the initial roll and pitch perturbations fairly quickly. 


\begin{figure*}[t]
    \centering
    \vspace{0.1in}
    \includegraphics[width = \linewidth]{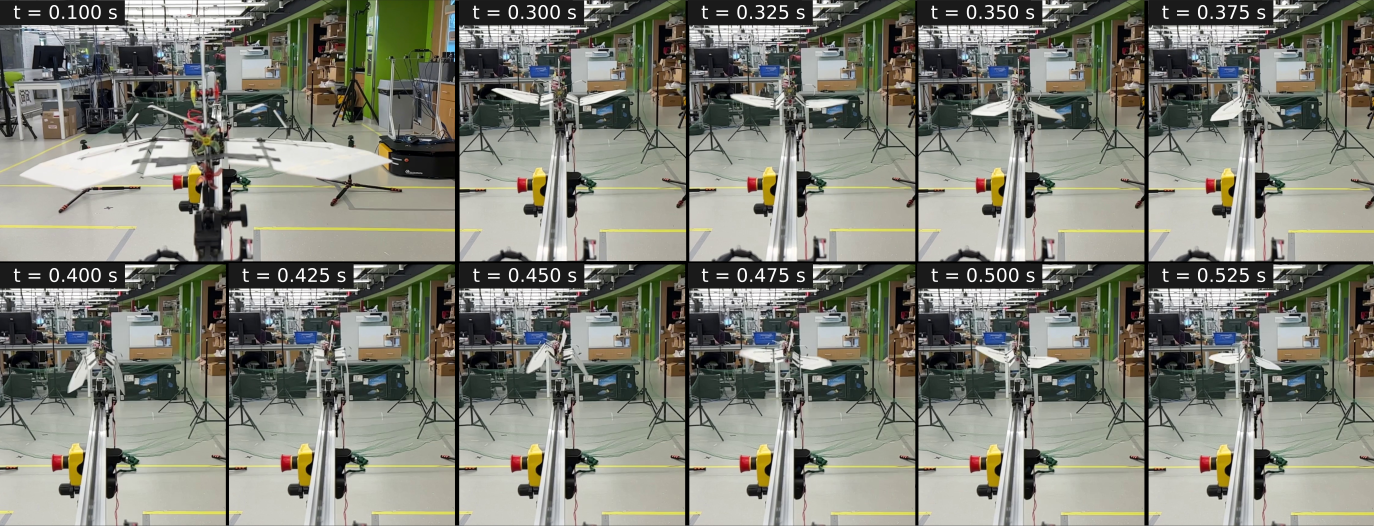}
    \caption{The snapshots of the Aerobat flight experiment which was captured using a high-speed camera showing the moment after launch and one flapping gait cycle. The robot was launched and flapping as it was being stabilized by the bang-bang control.}
    \vspace{-0.1in}
    \label{fig:experimental_results}
\end{figure*}

\begin{figure*}[t]
    \centering
    \vspace{0.1in}
    \includegraphics[width = \linewidth]{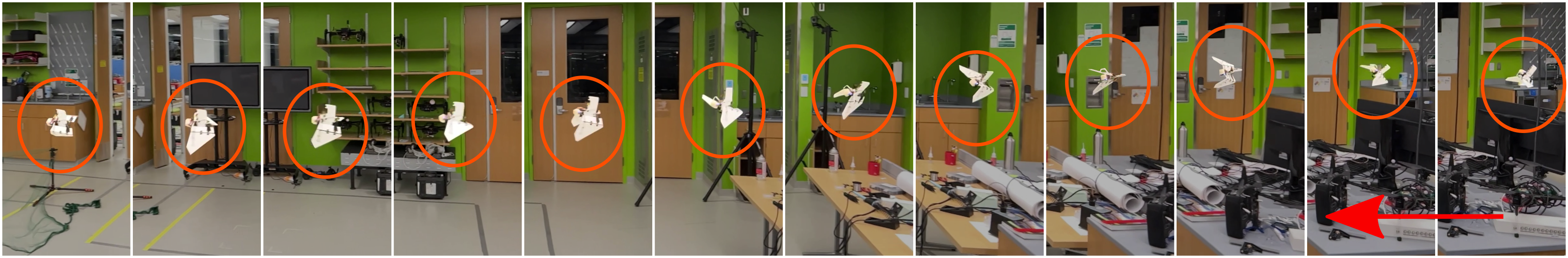}
    \caption{Illustrates flight trajectory after launch. Aerobat is launched from right and flies towards the net that is placed on the left-hand side. During the flight test, the bang-bang controller regulates the roll and pitch dynamics.}
    \vspace{-0.1in}
    \label{fig:flight-snapshots}
\end{figure*}

\begin{figure}[t]
    \centering
    \vspace{0.1in}
    \includegraphics[width = \linewidth]{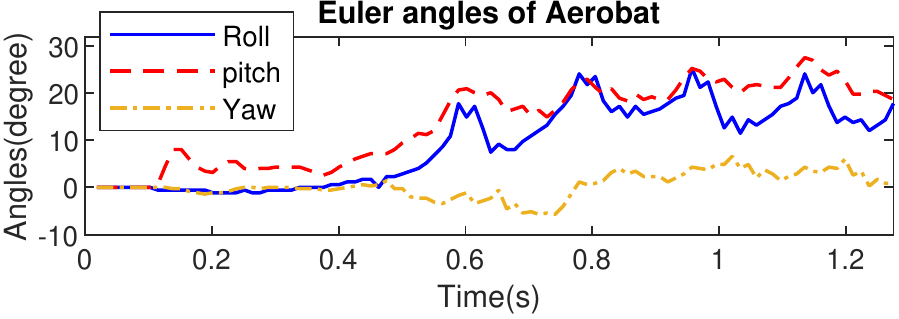}
    \caption{Shows Euler angles as measured by the onboard IMU during the flight experiment. The robot's orientation is relatively stable during the flight.}
    \vspace{-0.1in}
    \label{fig:experiment_plots}
\end{figure}

\subsection{Experiments}

To ensure a controlled takeoff of Aerobat, we design and build a motorized launcher, which can generate a consistent initial speed and orientation for Aerobat. This launcher can be rotated along the base on a tripod to launch Aerobat at different takeoff orientations. The launcher can achieve a maximum launch speed of 4 m/s precisely and consistently as it is actuated with a custom-made actuator controlled with a PID velocity controller. 

In our experiment, we set the controller's pitch and roll offset values to $20^\circ$ and $0^\circ$, respectively. We set Aerobat's flapping speed to approximately 5Hz, and the launch speed was set to be 2m/s to follow our simulation conditions. Figures \ref{fig:experimental_results} and \ref{fig:flight-snapshots} show the snapshots from our high-speed camera recordings of the flight experiment until a few moments before the robot landed onto a net. Figure \ref{fig:experiment_plots} shows the logged IMU data of Aerobat's orientation transmitted by an onboard Bluetooth system in the robot. 

As shown in Fig. \ref{fig:experiment_plots}, the launcher begins to accelerate at approximately 0.1s, and releases Aerobat at 0.3s. The launcher is designed to introduce a minimized interference in Aerobat's flight dynamics at the launch moment. However, external perturbations introduced by the launcher are inevitable. For instance, we observed a pitching moment imparted by the gripper upon release. After full release, Aerobat bang-bang controller is activated. 

According to our experimental results, the pitch and roll angles oscillate at a mean angle of approximately $20^\circ$ and $15^\circ$, respectively. The pitch and roll angles possess a peak-to-peak amplitude of $9^\circ$, respectively. We also noticed that the pitch angle oscillations are smaller compared to our simulation results, which could be attributed to a few known factors. These factors are not considered in our proposed modeling efforts include (1) lack of speed control in the KS mechanism, (2) asymmetry in the right and left wing flapping due to mechanical failure, (3) center-of-mass positions from model and actual robot do not exactly match, and (4) unknown drag force introduced by the thrusters and their impulsive behavior due to the bang-bang control scheme. 

For instance, the DC motor that drives the KS is controlled by a conventional sensor-less electronics speed controller which is incapable of maintaining a constant flapping speed in a single gait cycle. This issue causes flapping speed fluctuations during experiments which consequently manifest its effect in how the center-of-pressure translates relative to the body in a gait cycle.

Another important observation regarding model-experiment inconsistencies is about the yaw dynamics. Although the yaw dynamics is perfectly dormant in simulation due to inherent symmetry in the model, we observed yaw angle from experiments is affected by other factors that are not present in the model. For instance, the thrusters can generate a yaw moment due to aerodynamic reactive friction drag forces on the tiny propellers. 

\section{Conclusions and Future Work}
\label{sec:conclusion}

In this work, we present the dynamic modeling, simulation, and experimentation of a tail-less morphing flapping-wing robot which is stabilized using bang-bang control. The simulation shows the proof-of-concept of using the thrusters to help the robot stabilize its longitudinal and frontal dynamics, and the experimental results show that the robot is dynamically stable during the flight. For our future work, we will attempt to improve the stability of the flight and address the mechanical and control issues that are present in the current Aerobat build.


\printbibliography

@article{hubel_wake_2010,
	title = {Wake structure and wing kinematics: the flight of the lesser dog-faced fruit bat, {Cynopterus} brachyotis},
	volume = {213},
	issn = {0022-0949},
	shorttitle = {Wake structure and wing kinematics},
	url = {https://doi.org/10.1242/jeb.043257},
	doi = {10.1242/jeb.043257},
	number = {20},
	urldate = {2021-08-04},
	journal = {Journal of Experimental Biology},
	author = {Hubel, Tatjana Y. and Riskin, Daniel K. and Swartz, Sharon M. and Breuer, Kenneth S.},
	month = oct,
	year = {2010},
	pages = {3427--3440},
}

@article{sihite_integrated_2021,
	title = {An {Integrated} {Mechanical} {Intelligence} and {Control} {Approach} {Towards} {Flight} {Control} of {Aerobat}},
	url = {http://arxiv.org/abs/2103.16566},
	urldate = {2021-07-05},
	journal = {arXiv:2103.16566 [cs, eess]},
	author = {Sihite, Eric and Darabi, Atefe and Dangol, Pravin and Lessieur, Andrew and Ramezani, Alireza},
	month = mar,
	year = {2021},
	note = {arXiv: 2103.16566},
}

@misc{send_artificial_2012,
	title = {{ARTIFICIAL} {HINGED}-{WING} {BIRD} {WITH} {ACTIVE} {TORSION} {AND} {PARTIALLY} {LINEAR} {KINEMATICS}},
	url = {https://www.semanticscholar.org/paper/ARTIFICIAL-HINGED-WING-BIRD-WITH-ACTIVE-TORSION-AND-Send/cb886bb600682bda1062439cacbaf0eae3b2f865},
	language = {en},
	urldate = {2021-07-04},
	author = {Send, W.},
	year = {2012},
}

@article{wissa_free_2015,
	title = {Free {Flight} {Testing} and {Performance} {Evaluation} of a {Passively} {Morphing} {Ornithopter}},
	volume = {7},
	issn = {1756-8293},
	url = {https://doi.org/10.1260/1756-8293.7.1.21},
	doi = {10.1260/1756-8293.7.1.21},
	language = {en},
	number = {1},
	urldate = {2021-07-04},
	journal = {International Journal of Micro Air Vehicles},
	author = {Wissa, Aimy and Grauer, Jared and Guerreiro, Nelson and Hubbard, James and Altenbuchner, Cornelia and Tummala, Yashwanth and Frecker, Mary and Roberts, Richard},
	month = mar,
	year = {2015},
	note = {Publisher: SAGE Publications Ltd STM},
	pages = {21--40},
}

@article{gerdes_robo_2014,
	title = {Robo {Raven}: {A} {Flapping}-{Wing} {Air} {Vehicle} with {Highly} {Compliant} and {Independently} {Controlled} {Wings}},
	volume = {1},
	issn = {2169-5172},
	shorttitle = {Robo {Raven}},
	url = {https://www.liebertpub.com/doi/abs/10.1089/soro.2014.0019},
	doi = {10.1089/soro.2014.0019},
	number = {4},
	urldate = {2021-07-04},
	journal = {Soft Robotics},
	author = {Gerdes, John and Holness, Alex and Perez-Rosado, Ariel and Roberts, Luke and Greisinger, Adrian and Barnett, Eli and Kempny, Johannes and Lingam, Deepak and Yeh, Chen-Haur and Bruck, Hugh A. and Gupta, Satyandra K.},
	month = dec,
	year = {2014},
	note = {Publisher: Mary Ann Liebert, Inc., publishers},
	pages = {275--288},
}

@article{de_croon_design_2009,
	title = {Design, {Aerodynamics}, and {Vision}-{Based} {Control} of the {DelFly}},
	volume = {1},
	issn = {1756-8293},
	url = {https://doi.org/10.1260/175682909789498288},
	doi = {10.1260/175682909789498288},
	language = {en},
	number = {2},
	urldate = {2021-07-04},
	journal = {International Journal of Micro Air Vehicles},
	author = {de Croon, G.C.H.E. and de Clercq, K.M.E. and Ruijsink, R. and Remes, B. and de Wagter, C.},
	month = jun,
	year = {2009},
	note = {Publisher: SAGE Publications Ltd STM},
	pages = {71--97},
}

@inproceedings{rosen_development_2016,
	title = {Development of a 3.2g untethered flapping-wing platform for flight energetics and control experiments},
	doi = {10.1109/ICRA.2016.7487492},
	booktitle = {2016 {IEEE} {International} {Conference} on {Robotics} and {Automation} ({ICRA})},
	author = {Rosen, Michelle H. and le Pivain, Geoffroy and Sahai, Ranjana and Jafferis, Noah T. and Wood, Robert J.},
	month = may,
	year = {2016},
	pages = {3227--3233},
}

@article{ma_controlled_2013,
	title = {Controlled {Flight} of a {Biologically} {Inspired}, {Insect}-{Scale} {Robot}},
	volume = {340},
	copyright = {Copyright © 2013, American Association for the Advancement of Science},
	issn = {0036-8075, 1095-9203},
	url = {https://science.sciencemag.org/content/340/6132/603},
	doi = {10.1126/science.1231806},
	language = {en},
	number = {6132},
	urldate = {2021-07-04},
	journal = {Science},
	author = {Ma, Kevin Y. and Chirarattananon, Pakpong and Fuller, Sawyer B. and Wood, Robert J.},
	month = may,
	year = {2013},
	pmid = {23641114},
	note = {Publisher: American Association for the Advancement of Science
Section: Report},
	pages = {603--607},
}

@inproceedings{ramezani_bat_2016,
	address = {Stockholm, Sweden},
	title = {Bat {Bot} ({B2}), a biologically inspired flying machine},
	isbn = {978-1-4673-8026-3},
	url = {http://ieeexplore.ieee.org/document/7487491/},
	doi = {10.1109/ICRA.2016.7487491},
	language = {en},
	urldate = {2021-03-01},
	booktitle = {2016 {IEEE} {International} {Conference} on {Robotics} and {Automation} ({ICRA})},
	publisher = {IEEE},
	author = {Ramezani, Alireza and Shi, Xichen and Chung, Soon-Jo and Hutchinson, Seth},
	month = may,
	year = {2016},
	pages = {3219--3226},
}

@incollection{mangan_describing_2017,
	address = {Cham},
	title = {Describing {Robotic} {Bat} {Flight} with {Stable} {Periodic} {Orbits}},
	volume = {10384},
	isbn = {978-3-319-63536-1 978-3-319-63537-8},
	url = {http://link.springer.com/10.1007/978-3-319-63537-8_33},
	language = {en},
	urldate = {2021-03-01},
	booktitle = {Biomimetic and {Biohybrid} {Systems}},
	publisher = {Springer International Publishing},
	author = {Ramezani, Alireza and Ahmed, Syed Usman and Hoff, Jonathan and Chung, Soon-Jo and Hutchinson, Seth},
	editor = {Mangan, Michael and Cutkosky, Mark and Mura, Anna and Verschure, Paul F.M.J. and Prescott, Tony and Lepora, Nathan},
	year = {2017},
	doi = {10.1007/978-3-319-63537-8_33},
	note = {Series Title: Lecture Notes in Computer Science},
	pages = {394--405},
}

@inproceedings{hoff_trajectory_2019,
	address = {Macau, China},
	title = {Trajectory planning for a bat-like flapping wing robot},
	isbn = {978-1-72814-004-9},
	url = {https://ieeexplore.ieee.org/document/8968450/},
	doi = {10.1109/IROS40897.2019.8968450},
	language = {en},
	urldate = {2021-03-01},
	booktitle = {2019 {IEEE}/{RSJ} {International} {Conference} on {Intelligent} {Robots} and {Systems} ({IROS})},
	publisher = {IEEE},
	author = {Hoff, Jonathan and Syed, Usman and Ramezani, Alireza and Hutchinson, Seth},
	month = nov,
	year = {2019},
	pages = {6800--6805},
}

@inproceedings{hoff_synergistic_2016,
	title = {Synergistic {Design} of a {Bio}-{Inspired} {Micro} {Aerial} {Vehicle} with {Articulated} {Wings}},
	isbn = {978-0-9923747-2-3},
	url = {http://www.roboticsproceedings.org/rss12/p09.pdf},
	doi = {10.15607/RSS.2016.XII.009},
	language = {en},
	urldate = {2021-03-01},
	booktitle = {Robotics: {Science} and {Systems} {XII}},
	publisher = {Robotics: Science and Systems Foundation},
	author = {Hoff, Jonathan and Ramezani, Alireza and Chung, Soon-Jo and Hutchinson, Seth},
	year = {2016},
}

@inproceedings{ramezani_lagrangian_2015,
	address = {Hamburg, Germany},
	title = {Lagrangian modeling and flight control of articulated-winged bat robot},
	isbn = {978-1-4799-9994-1},
	url = {http://ieeexplore.ieee.org/document/7353772/},
	doi = {10.1109/IROS.2015.7353772},
	language = {en},
	urldate = {2021-03-01},
	booktitle = {2015 {IEEE}/{RSJ} {International} {Conference} on {Intelligent} {Robots} and {Systems} ({IROS})},
	publisher = {IEEE},
	author = {Ramezani, Alireza and Shi, Xichen and Chung, Soon-Jo and Hutchinson, Seth},
	month = sep,
	year = {2015},
	pages = {2867--2874},
}

@article{ramezani_biomimetic_2017,
	title = {A biomimetic robotic platform to study flight specializations of bats},
	volume = {2},
	issn = {2470-9476},
	url = {https://robotics.sciencemag.org/lookup/doi/10.1126/scirobotics.aal2505},
	doi = {10.1126/scirobotics.aal2505},
	language = {en},
	number = {3},
	urldate = {2021-03-01},
	journal = {Science Robotics},
	author = {Ramezani, Alireza and Chung, Soon-Jo and Hutchinson, Seth},
	month = feb,
	year = {2017},
	pages = {eaal2505},
}

@incollection{mangan_reducing_2017,
	address = {Cham},
	title = {Reducing {Versatile} {Bat} {Wing} {Conformations} to a 1-{DoF} {Machine}},
	volume = {10384},
	isbn = {978-3-319-63536-1 978-3-319-63537-8},
	url = {http://link.springer.com/10.1007/978-3-319-63537-8_16},
	language = {en},
	urldate = {2021-03-01},
	booktitle = {Biomimetic and {Biohybrid} {Systems}},
	publisher = {Springer International Publishing},
	author = {Hoff, Jonathan and Ramezani, Alireza and Chung, Soon-Jo and Hutchinson, Seth},
	editor = {Mangan, Michael and Cutkosky, Mark and Mura, Anna and Verschure, Paul F.M.J. and Prescott, Tony and Lepora, Nathan},
	year = {2017},
	doi = {10.1007/978-3-319-63537-8_16},
	note = {Series Title: Lecture Notes in Computer Science},
	pages = {181--192},
}

@inproceedings{sihite_enforcing_2020,
	address = {Jeju Island, Korea (South)},
	title = {Enforcing nonholonomic constraints in {Aerobat}, a roosting flapping wing model},
	isbn = {978-1-72817-447-1},
	url = {https://ieeexplore.ieee.org/document/9304158/},
	doi = {10.1109/CDC42340.2020.9304158},
	language = {en},
	urldate = {2021-03-01},
	booktitle = {2020 59th {IEEE} {Conference} on {Decision} and {Control} ({CDC})},
	publisher = {IEEE},
	author = {Sihite, Eric and Ramezani, Alireza},
	month = dec,
	year = {2020},
	pages = {5321--5327},
}

@article{sihite_computational_2020,
	title = {Computational {Structure} {Design} of a {Bio}-{Inspired} {Armwing} {Mechanism}},
	volume = {5},
	issn = {2377-3766, 2377-3774},
	url = {https://ieeexplore.ieee.org/document/9143405/},
	doi = {10.1109/LRA.2020.3010217},
	language = {en},
	number = {4},
	urldate = {2021-03-01},
	journal = {IEEE Robotics and Automation Letters},
	author = {Sihite, Eric and Kelly, Peter and Ramezani, Alireza},
	month = oct,
	year = {2020},
	pages = {5929--5936},
}

@article{hoff_optimizing_2018,
	title = {Optimizing the structure and movement of a robotic bat with biological kinematic synergies},
	volume = {37},
	issn = {0278-3649, 1741-3176},
	url = {http://journals.sagepub.com/doi/10.1177/0278364918804654},
	doi = {10.1177/0278364918804654},
	language = {en},
	number = {10},
	urldate = {2021-03-01},
	journal = {The International Journal of Robotics Research},
	author = {Hoff, Jonathan and Ramezani, Alireza and Chung, Soon-Jo and Hutchinson, Seth},
	month = sep,
	year = {2018},
	pages = {1233--1252},
}

@article{riskin_bats_2009,
	title = {Bats go head-under-heels: the biomechanics of landing on a ceiling.},
	volume = {212},
	shorttitle = {Bats go head-under-heels},
	number = {Pt},
	journal = {The Journal of experimental biology},
	author = {Riskin, Daniel K. and Bahlman, Joseph W. and Hubel, Tatjana Y. and Ratcliffe, John M. and Kunz, Thomas Hans and Swartz, Sharon M.},
	year = {2009},
	pages = {945--953},
}

@article{iriarte-diaz_whole-body_2011,
	title = {Whole-body kinematics of a fruit bat reveal the influence of wing inertia on body accelerations},
	volume = {214},
	language = {en},
	number = {9},
	urldate = {2019-09-04},
	journal = {Journal of Experimental Biology},
	author = {Iriarte-Diaz, J. and Riskin, D. K. and Willis, D. J. and Breuer, K. S. and Swartz, S. M.},
	month = may,
	year = {2011},
	pages = {1546--1553},
}

@article{farrell_helbling_review_2018,
	title = {A {Review} of {Propulsion}, {Power}, and {Control} {Architectures} for {Insect}-{Scale} {Flapping}-{Wing} {Vehicles}},
	volume = {70},
	language = {en},
	number = {1},
	urldate = {2020-05-22},
	journal = {Applied Mechanics Reviews},
	author = {Farrell Helbling, E. and Wood, Robert J.},
	month = jan,
	year = {2018},
}

@article{boutet_unsteady_2018,
	title = {Unsteady {Lifting} {Line} {Theory} {Using} the {Wagner} {Function} for the {Aerodynamic} and {Aeroelastic} {Modeling} of {3D} {Wings}},
	volume = {5},
	copyright = {http://creativecommons.org/licenses/by/3.0/},
	language = {en},
	number = {3},
	urldate = {2020-12-31},
	journal = {Aerospace},
	author = {Boutet, Johan and Dimitriadis, Grigorios},
	month = sep,
	year = {2018},
	pages = {92},
}

@article{riskin_upstroke_2012,
	title = {Upstroke wing flexion and the inertial cost of bat flight},
	volume = {279},
	language = {eng},
	number = {1740},
	journal = {Proceedings. Biological Sciences},
	author = {Riskin, Daniel K. and Bergou, Attila and Breuer, Kenneth S. and Swartz, Sharon M.},
	month = aug,
	year = {2012},
	pages = {2945--2950},
}

@article{phan_insect-inspired_2019,
	title = {Insect-inspired, tailless, hover-capable flapping-wing robots: {Recent} progress, challenges, and future directions},
	volume = {111},
	issn = {0376-0421},
	shorttitle = {Insect-inspired, tailless, hover-capable flapping-wing robots},
	url = {http://www.sciencedirect.com/science/article/pii/S0376042119300545},
	doi = {10.1016/j.paerosci.2019.100573},
	language = {en},
	urldate = {2020-12-04},
	journal = {Progress in Aerospace Sciences},
	author = {Phan, Hoang Vu and Park, Hoon Cheol},
	month = nov,
	year = {2019},
	pages = {100573},
}

@article{sane_lift_2001,
	title = {Lift and drag production by a flapping wing},
	language = {en},
	author = {Sane, S P and Dickinson, M H},
	year = {2001},
	pages = {20}
}

@article{peterson_wing-assisted_2011,
	title = {A wing-assisted running robot and implications for avian flight evolution},
	volume = {6},
	issn = {1748-3182, 1748-3190},
	url = {https://iopscience.iop.org/article/10.1088/1748-3182/6/4/046008},
	doi = {10.1088/1748-3182/6/4/046008},
	language = {en},
	number = {4},
	urldate = {2021-03-01},
	journal = {Bioinspir. Biomim.},
	author = {Peterson, K and Birkmeyer, P and Dudley, R and Fearing, R S},
	month = dec,
	year = {2011},
	pages = {046008}
}

@article{chukewad_robofly_2020,
	title = {{RoboFly}: {An} insect-sized robot with simplified fabrication that is capable of flight, ground, and water surface locomotion},
	shorttitle = {{RoboFly}},
	url = {http://arxiv.org/abs/2001.02320},
	language = {en},
	urldate = {2021-03-01},
	journal = {arXiv:2001.02320 [cs, eess]},
	author = {Chukewad, Yogesh M. and James, Johannes and Singh, Avinash and Fuller, Sawyer},
	month = oct,
	year = {2020},
	note = {arXiv: 2001.02320}
}

@article{tu_untethered_2020,
	title = {Untethered {Flight} of an {At}-{Scale} {Dual}-motor {Hummingbird} {Robot} with {Bio}-inspired {Decoupled} {Wings}},
	issn = {2377-3766, 2377-3774},
	url = {https://ieeexplore.ieee.org/document/9001181/},
	doi = {10.1109/LRA.2020.2974717},
	language = {en},
	urldate = {2021-03-02},
	journal = {IEEE Robot. Autom. Lett.},
	author = {Tu, Zhan and Fei, Fan and Deng, Xinyan},
	year = {2020},
	pages = {1--1}
}

@article{hedenstrom_bat_2015,
	title = {Bat flight: aerodynamics, kinematics and flight morphology},
	volume = {218},
	issn = {0022-0949, 1477-9145},
	url = {http://jeb.biologists.org/cgi/doi/10.1242/jeb.031203},
	doi = {10.1242/jeb.031203},
	shorttitle = {Bat flight},
	pages = {653--663},
	number = {5},
	journaltitle = {Journal of Experimental Biology},
	author = {Hedenstrom, A. and Johansson, L. C.},
	urldate = {2020-11-24},
	date = {2015-03-01},
	langid = {english}
}


\end{document}